\documentclass[runningheads]{llncs}
\usepackage[T1]{fontenc}
\usepackage{graphicx,verbatim}
\usepackage[misc,geometry]{ifsym}
\usepackage{booktabs}
\usepackage{subcaption}
\usepackage{xcolor}
\usepackage{amsmath}
\usepackage{cleveref}
\usepackage{array}
\usepackage{tabularx} 
\usepackage{graphicx}
\def\eg{\emph{e.g.}} 
\def\ie{\emph{i.e.}}

\def\etal{\emph{et al.}}
\usepackage{amsfonts}

\newcolumntype{P}[1]{>{\centering\arraybackslash}p{#1}}
\newcolumntype{M}[1]{>{\centering\arraybackslash}m{#1}}

\usepackage{xcolor}
\definecolor{red}{RGB}{230,10,30}

\begin{document}
\title{Conquering the Retina: \\ Bringing Visual in-Context Learning to OCT}
%

\author{Alessio Negrini \and
Simon Rei{\ss}} 
\authorrunning{A. Negrini and S. Rei{\ss}}
%
\institute{Karlsruhe Institute of Technology, 76131 Karlsruhe, Germany
\Letter~\email{simon.reiss@kit.edu}}


\maketitle              
\begin{abstract}
Recent advancements in medical image analysis have led to the development of highly specialized models tailored to specific clinical tasks.
These models have demonstrated exceptional performance and remain a crucial research direction.
Yet, their applicability is limited to predefined tasks, requiring expertise and extensive resources for development and adaptation.
In contrast, generalist models offer a different form of utility: allowing medical practitioners to define tasks on the fly without the need for task-specific model development.
In this work, we explore how to train generalist models for the domain of retinal optical coherence tomography using visual in-context learning (VICL), \ie, training models to generalize across tasks based on a few examples provided at inference time.
To facilitate rigorous assessment, we propose a broad evaluation protocol tailored to VICL in OCT.
We extensively evaluate a state-of-the-art medical VICL approach on multiple retinal OCT datasets, establishing a first baseline to highlight the potential and current limitations of in-context learning for OCT.
To foster further research and practical adoption, we openly release our code~\footnote{\url{https://github.com/negralessio/thesis-visual-in-context-learning}}.


\end{abstract}

\section{Introduction}
Medical image analysis has been pushed forward significantly through the adaptation of deep neural nets~\cite{ronneberger2015u,isensee2021nnu,huang2023stu}.
Through the curation of datasets, the successive annotation by medical experts and supervised training of neural models, tasks such as classification-, segmentation- or even style-transfer tasks can be addressed with increasing accuracy.
While this progress is leading to solutions which fulfill the high standards of medical applications, each new task to be solved at the same time requires going through the costly steps of data collection, annotation and model training.
To circumvent this, some works in literature propose to design the neural architectures and training strategies such that they are extensible to new contexts, \ie, new tasks to be addressed at test time without model re-training~\cite{czolbe2023neuralizer,ren2024medical}, so called visual in-context learning models.
This shift enables medical practitioners to solve highly individual use-cases by providing new context information at test time.

The design of architectures and training objectives for visual in-context learning are currently heavily researched open questions~\cite{czolbe2023neuralizer,ren2024medical,bai2024sequential,wang2023images,bar2022visual}.
In our work, we start by exploring the data- and annotation requirements for training, \ie, whether a set of images and segmentation annotations are a suitable enough starting point to train an in-context learning model.
Specifically, we investigate this in the medical image domain of retinal optical coherence tomography (OCT).
We then setup an evaluation protocol to benchmark trained visual in-context learning models in their efficacy in adapting to new, unseen tasks without re-training and we investigate the boundaries of the model's transfer capabilities by testing the adaptation to unseen tasks \textit{and} unseen data distributions, so called domain generalization.
In our exploration, we heavily utilize the Neuralizer architecture~\cite{czolbe2023neuralizer} and train it on retinal OCTs leading to our \emph{Retinalizer} models.
We further enable the models to draw multi-class predictions and explore the effect of a simple and effective \emph{random recoloring task-augmentation} to improve the generalization capabilities of trained in-context learners.
Our extensive -- and to our best knowledge -- first exploration into visual in-context learning for OCT allows a first glimpse of the potential of generalist models in the domain of optical coherence tomography.
Our contributions summarize to:
\begin{itemize}
    \item We investigate visual in-context learning for retinal optical coherence tomography for the first time by using image- and segmentation data only.
    \item We propose a design to draw multi-class predictions from a medical visual in-context learning model and a color-based task-augmentation technique.
    \item We extensively evaluate generalization efficacy of trained models to unseen tasks and unseen data distributions in optical coherence tomography.
\end{itemize}

\section{Visual in-Context Learning as Conditional Prediction}

In visual in-context learning, a model $\theta(\cdot, \cdot)$ is trained, which based on a task-specific context set $C$ can transform a query image $Q$ into a task-specific output via $\theta(C, Q) = O$.
As such, in case the context set $C$ conditions the model to the task of segmentation, the model should yield a segmentation of $Q$.
To inform the model $\theta$ how $Q$ has to be transformed, visual in-context learning utilizes example images of the task to be solved in $C$.
For convenient and coherent provision of this task information to the model, all tasks are formulated as image-to-image tasks (\ie, in- and output are both RGB images), which leads to a context set $C \in \mathbb{R}^{n \times 2 \times 3 \times width \times height}$, where $n$ specifies the number of image-pair examples to describe the task.
This formulation does not only allow tasks such as denoising or inpainting, which naturally are image-to-image tasks, but also tasks like semantic segmentation: Instead of predicting a 1-hot class vector for each pixel the model predicts a color value, with class-color associations defined in the image-pairs of the context set $C$.

This unified interface of defining all tasks in terms of image-to-image processing enables predictions with context sets $C^{unseen}$, which contain tasks not observed in training.
As such, the predictive capabilities of $\theta$ to generalize to unseen context sets is the target measure to evaluate in visual in-context learning.

\section{From Neuralizer to Retinalizer}
Training a visual in-context learning model that can generalize to new tasks generally requires a large dataset containing many tasks to pre-train on~\cite{czolbe2023neuralizer,bai2024sequential}. 
As vessel for medical visual in-context learning, we utilize the network architecture Neuralizer~\cite{czolbe2023neuralizer}.
Neuralizer allows for in-context learning by introducing Pairwise-Conv-Avg blocks, which enable processing context sets of varying size $n$ into a Unet architecture~\cite{ronneberger2015u}.
As the name suggests, Neuralizer is originally trained on neuro images.
Next, we outline our data pre-processing and training strategy to bring such a visual in-context learning model to the domain of optical coherence tomography, which due to the focus on the retina in OCT, we term \emph{Retinalizer}.

\subsection{Base Optical Coherence Tomography Datasets}
\label{sec:datasets}
To get hold of enough data to train visual in-context learners, we make use of publicly available datasets from the OCT community, namely the DUKE dataset family for cyst- and retinal layer segmentation, the UMN dataset with diabetic macular edema fluid annotations, the RETOUCH dataset family for multi fluid segmentation with its diversity of three OCT device vendors and the OCTDL dataset, which covers OCT scans of diverse medical conditions in the retina.

\textbf{DUKE}~\cite{chiu2015kernel} contains OCT scans which are annotated with diabetic macular edema (DME) fluids and, for a second set of scans also with the different anatomical layers of the retina.
DME labeled scans amount to $610$ of which $532$ contain healthy retinas, while the layer segmentation portion includes $110$ scans.

\textbf{UMN}~\cite{rashno2017fully} also contains pixel-wise annotations of fluid accumulations in patients with DME.
However, it has no severe class imbalance as it contains $725$ samples in total, taken from $29$ patients with $25$ scans per subject.
It includes $405$ segmentation masks of which ca. $56\%$ depict healthy retinal sections.

\textbf{RETOUCH}~\cite{bogunovic2019retouch} contains scans with manual annotations for retinal fluids acquired with devices of three different OCT manufactures: Topcon, Cirrus and Spectralis.
Retinal fluids are subdivided into Intra-retinal fluid, Sub-retinal fluid and Pigment Epithelial Detachment.
Of the $6,936$ scans in the dataset $3,563$ (ca. $50\%$) are healthy, in the remaining scans $2,068$ (ca. $30\%$) contain one fluid class, $1,013$ (ca. $15\%$) contain two and $292$ (ca. $5\%$) contain all three classes.

\textbf{OCTDL}~\cite{kulyabin2024octdl} consists of around $1,618$ scans with ca. $16\%$ showing healthy retinas, while the rest show manifestations of diverse diseases.
As the scans come with classification labels, in our study, we use it to represent the diversity in OCT without directly utilizing the label information.
Specifically, OCTDL builds the basis of image-content tasks, where in- and output images depict retinal scans.

\subsection{Heterogeneous Task Enrichment}
\label{sec:tasks}
In addition to a large set of images, visual in-context training requires a diverse set of pre-training tasks exceeding merely segmenting retinal layers and fluids.
To enable training on such a task diversity, we propose to enrich the data by deriving additional tasks: semantic-, transformation- and generative tasks.

\textbf{Semantic tasks} are based on the binary- and multi-class segmentation annotations of the base OCT datasets.
As such, we pre-compute the tasks \textit{semantic edge detection}, where semantic regions are delimited by thin lines, \textit{semantic skeletons}, which indicate the medial axes of segments and \textit{coarse semantic segmentation}, which encompasses the semantic hull of each segment as crude approximate segmentation.
We also use the original \textit{segmentation tasks}, which leads to four \textit{semantic task types} for four pixel-wise labeled datasets, \ie, $16$ semantic tasks.

\textbf{Transformation tasks} are based on OCTDL, where a single scan serves as basis for both input and output.
We setup four transformation tasks, the first two are rotation tasks where the input image is transformed with a \textit{rotation} by either $90^\circ$ or $270^\circ$, the rotated image serves as output.
Similarly, in the \textit{image inversion task}, the intensity values $v$ of the input scan are inverted by $255 - v$, yielding the corresponding output.
The \textit{image revert task} is its counterpart, where the input and output are switched.
In total, this amounts to four additional tasks.

\begin{figure}[t]
    \centering
    \includegraphics[width=\textwidth,height=0.4\textwidth]{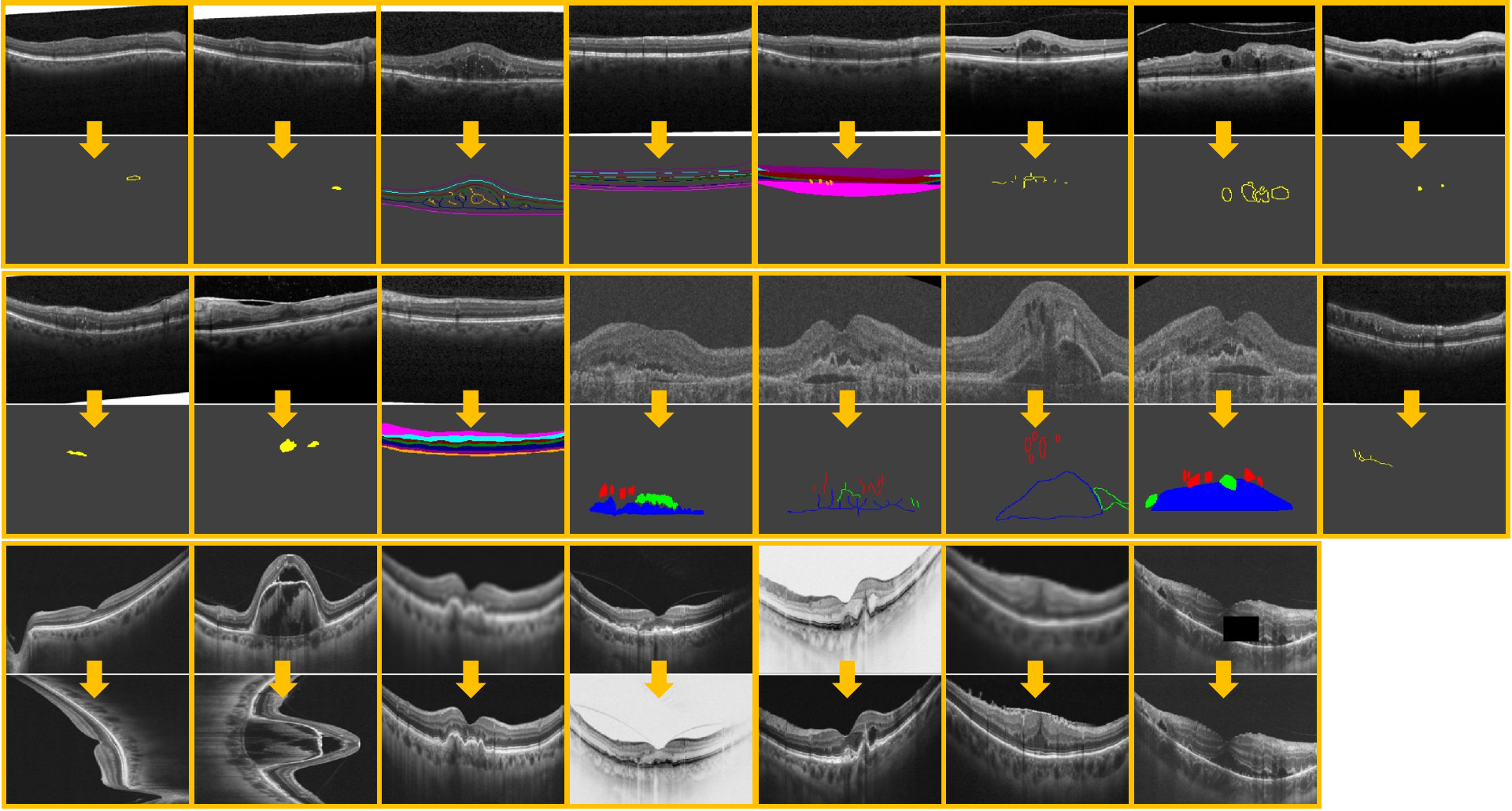}
    \caption{Examples for seen tasks. Input-output pairs grouped into orange boxes.}
    \label{fig:task_data}
\end{figure}

\textbf{Generative tasks} are based on OCTDL as well, but center around reconstructing the original scan from manipulated input scans.
One such task is \textit{Gaussian denoising} where the input is a heavily noised version of the output scan.
In the \textit{image super resolution} task, a scan is down-scaled to a lower resolution which is taken as manipulated input.
The task aims to reconstruct information from the low resolution image to the original, higher resolution image.
Finally, the \textit{image in-painting} task takes an image with a masked region as input, the original image is the output.
Thus, in total we have $23$ tasks as seen in~\Cref{fig:task_data}.

\subsection{Visual in-Context Training for Optical Coherence Tomography}
\label{sec:strategy}
With datasets and a broad set of pre-training tasks in place, the visual in-context model can be trained.
Accommodating the properties of OCT datasets and tasks, we adapted key components in the training strategy, which we describe next.

\textbf{Batch Sampling and Optimization}
To ensure, that the model does not predominantly focus on tasks which have a high number of samples during training, we utilize task balancing, \ie, tasks with fewer samples are over-sampled, while populous tasks are under-sampled.
This is necessary due to the imbalance in sample size of the tasks, which itself stems from our utilization of multiple datasets.
Further, in optimization we deviate from the Neuralizer training strategy to define task-specific loss functions~\cite{czolbe2023neuralizer} and instead reduce complexity during training by using a single reconstruction loss function:
\begin{equation}
    \mathcal{L}_{\text{MSE}} = \frac{1}{|B|} \sum_{i=1}^{|B|} \| y_i - \theta(C_i, Q_i) \|^2_2 \enspace,
\end{equation}
with the mini-batch $B$, $y_i$ indicating the ground-truth at batch index $i$ for the input $Q_i$ under the task-specific context set $C_i$ in the batch.

\textbf{Enabling multi-class predictions}
Out-of-the-box, the Neuralizer architecture does not allow for multi-class predictions, as it was designed for binary semantic prediction.
To address this, we define multi-class problems, \eg, segmentation of multiple types of retinal fluid by predicting differently colored segments.
To map the network prediction for semantic tasks back to the original classes, we extract all unique colors $v_c$ from the context set and compute the euclidean distance to each individual pixel-prediction $v_p$:
\begin{equation}
    d(v_p,\,v_c)=||v_p-v_c||_2 \;\;\;\;\; \forall v_c \in C \enspace.
\end{equation}
Each pixel vector $v_p$ in the prediction is then replaced with the color $v_c$ from the context set that has the lowest distance: $v_p \doteq arg min_{v_c} \; d(v_p, \,v_c) \;\;\forall v_c \in C$.

\textbf{Random recoloring augmentation for task-adaptiveness}
A high diversity in tasks during training is key for visual in-context learning.
Therefore, as in standard image processing, where diversity and quantity of training samples are increased through image augmentations such as rotation, cropping or color jitter, we aim to explore the effect of a task-augmentation for visual in-context learning.
Specifically, we holistically augment the context set $C$ and expected output $O$.
To this end, we apply a random recoloring augmentation on all semantic tasks, where coherently in $C$ and $O$, we randomly draw new colors for each present class and substitute the new class colors on the fly during training.

\subsection{A Visual in-Context Learning Testbed for OCT}
To get a clear view on the capabilities of visual in-context learners a clear evaluation protocol is key.
In this section we outline quantitative evaluation scenarios and baselines to rigorously measure model efficacy for previously unseen tasks.

\textbf{Evaluation of Task Adaptation Capabilities}
To investigate how well a visual in-context model is able to adapt to new tasks, \ie, tasks not present in training, we define additional unseen tasks.
For a broad investigation of how different such unseen tasks can be with respect to seen tasks, they are created to have varying degrees of discrepancies with them.
The tasks we define are \textit{binary layer segmentation}, where all layers of the DUKE dataset are merged into a single layer to be segmented, \textit{retinal boundary detection}, which we define as task to detect the outermost boundaries of the retina region and finally the task \textit{random recolored multi fluid segmentation}, where we chose randomly new colors for each fluid type.
We further define the tasks of \textit{salt and pepper denoising}, \ie, a new type of noise to be removed, \textit{$2\times$ in-painting}, where we task the model to in-paint two patches in an image and \textit{out-painting} where the model needs to fill a masked out frame at the border of the image.

In addition, on the RETOUCH dataset, we further evaluate the generalization capability of visual in-context models to new data distributions by training on data from two vendors and exclusively evaluate on data of the third, held out vendor in testing.
In these so called domain generalization experiments, the model needs to adapt to new unseen tasks and to new data distributions without information about them at training time.
These experiments are three-fold as we hold out data from the vendors Topcon, Cirrus and Spectralis, successively.

We split all data into train, val and test sets ($60/20/20$).
The VICL models are developed on train and val using the seen tasks, results are reported on unseen tasks in the test set.
As metrics, we report mean Intersection over Union (IoU) for segmentation, for sparse semantic structures such as boundaries we use the F-1 Score, generative tasks are evaluated with Mean Average Error (MAE).

{
\begin{table}[t]
    \setlength{\tabcolsep}{1pt}
    \scriptsize
    \caption{Generalization performance for models solving previously unseen tasks.}
    \centering
    \begin{tabular}{lc|ccccc}
        \toprule
        Task & Metric & Copy  & Neuralizer & Retinalizer & Retinalizer Rec. & Single-task \\
        \midrule
        D-Layer Bin. Seg. & IoU $\uparrow$ & $67.74 \pm 18.0$ & $41.43 \pm \phantom{0}1.0$ & $62.99 \pm \phantom{0}3.9$ & $\textbf{81.94} \pm \phantom{0}\textbf{4.9}$ & $\color{gray}{98.15 \pm \phantom{0}0.7}$ \\
        R-Rec. Fluid Seg. & IoU $\uparrow$ & $31.84 \pm 10.1$ & $\phantom{0}5.81 \pm 12.6$ & $38.53 \pm 13.1$ & $\textbf{53.64} \pm \textbf{15.7}$ & $\color{gray}{46.73 \pm 15.8}$ \\
        D-Layer Boundary & F-1 $\uparrow$ & $51.88 \pm \phantom{0}2.9$ & $49.73 \pm \phantom{0}0.0$ & $53.46 \pm \phantom{0}3.1 $ & $\textbf{60.69} \pm \phantom{0}\textbf{1.9}$ & $\color{gray}{65.58 \pm \phantom{0}6.9}$ \\
        O-$2\times$ in-painting & MAE $\downarrow$ & $0.143 \pm 0.026$ & $0.305 \pm 0.0360$ & $0.010 \pm 0.004$ & $\textbf{0.008} \pm \textbf{0.003}$ & $\color{gray}{0.007 \pm 0.003}$ \\
        R-$2\times$ in-painting & MAE $\downarrow$ & $0.115 \pm 0.031$ & $0.293 \pm 0.047$ & $0.013 \pm 0.005$ & $\textbf{0.009} \pm \textbf{0.005}$ & $\color{gray}{0.004 \pm 0.002}$ \\
        O-Outpainting & MAE $\downarrow$ & $0.143 \pm 0.025$ & $0.304 \pm 0.036$ & $0.014 \pm 0.003$ & $\textbf{0.013} \pm \textbf{0.003}$ & $\color{gray}{0.009 \pm 0.002}$ \\
        R-Outpainting & MAE $\downarrow$ & $0.113 \pm 0.030$ & $0.294 \pm 0.047$ & $0.017 \pm 0.005$ & $\textbf{0.014} \pm \textbf{0.004}$ & $\color{gray}{0.006 \pm 0.001}$ \\
        O-S\&P Denoising & MAE $\downarrow$ & $0.143 \pm 0.027$ & $0.305 \pm 0.037$ & $\textbf{0.017} \pm \textbf{0.004}$ & $0.028 \pm 0.007$ & $\color{gray}{0.001 \pm 0.001}$ \\
        R-S\&P Denoising & MAE $\downarrow$ & $0.114 \pm 0.030$ & $0.292 \pm 0.047$ & $\textbf{0.022} \pm \textbf{0.018}$ & $0.037 \pm 0.020$ & $\color{gray}{0.001 \pm 0.000}$ \\
    \bottomrule
    \end{tabular}
    \label{tab:unseen}
\end{table}
}
\textbf{Baselines}
Here, we briefly outline the baseline methods we compare to when evaluating visual in-context learning models in the above scenarios.

\noindent \underline{Copy} chooses one of the output images from the context set $C$ and returns it as `prediction'.
This is a naive lower bound that trained models should exceed.

\noindent \underline{Neuralizer} is a model by Czolbe~\etal, trained on neural imaging and evaluated on OCT data.
Models trained in the OCT domain should outperform this model.

\noindent \underline{Retinalizer} is our re-trained model using the same architecture as Neuralizer, but trained on OCT as outlined in~\Cref{sec:datasets},~\Cref{sec:tasks} and~\Cref{sec:strategy}.

\noindent \underline{Retinalizer Rec.} is a similarly trained model as Retinalizer, with the distinction, that random recoloring augmentation is enabled during training.

\noindent \underline{Single-task} baselines are multiple models with the architecture of Neuralizer, but trained in a supervised fashion on each task individually.
As such each model was trained on the respective unseen tasks and thus presents a loose upper bound. \\

\noindent \textbf{Implementation details}
We train the Retinalizer models on $3 \times 192 \times 192$ images, with batch size $5$ and context set size $|C| = 6$.
We further alter the sampling process of the context set for semantic tasks such that at least $2$ of the image-pairs include non-empty masks.
We use a learning rate of $0.0002$, Adam~\cite{kingma2014adam} with $(\beta_1,\beta_2) = (0.5, 0.999)$ and train $16$ epochs on a RTX 2080 Ti.

\section{Experimental Results}


{
\setlength{\tabcolsep}{3pt}
\begin{table}[t]
    \caption{Domain generalization results on unseen RETOUCH tasks.}
    \centering
    \scriptsize
    \begin{tabular}{lc|cccc}
    \toprule
        Task & Metric & Copy & Retinalizer & Retinalizer Rec. & Single-task \\
        \midrule

        & & \multicolumn{4}{c}{\textit{Domain generalization to RETOUCH-spectralis}} \\
        R-Rec. Fluid Seg. & IoU $\uparrow$ & $31.09 \pm \phantom{0}9.6$ & $31.89 \pm 10.1$ & $\textbf{39.80} \pm \textbf{10.4}$ & $\color{gray}{37.56 \pm 11.4}$ \\
        R-$2\times$ in-painting & MAE $\downarrow$ & $0.144 \pm 0.030 $ & $0.014 \pm 0.004$ & $\textbf{0.013} \pm \textbf{0.004}$ & $\color{gray}{0.007 \pm 0.002}$ \\
        R-Outpainting & MAE $\downarrow$ & $0.143 \pm 0.029$ & $0.020 \pm 0.003$ & $\textbf{0.019} \pm \textbf{0.004}$ & $\color{gray}{0.010 \pm 0.002}$ \\
        R-S\&P Denoising & MAE $\downarrow$ & $0.144 \pm 0.028$ & $0.036 \pm 0.012$ & $\textbf{0.031} \pm \textbf{0.011}$ & $\color{gray}{0.002 \pm 0.000}$ \\

        & & \multicolumn{4}{c}{\textit{Domain generalization to RETOUCH-topcon}} \\
        R-Rec. Fluid Seg. & IoU $\uparrow$ & $35.25 \pm 12.0 $ & $38.75 \pm 11.4 $ & $\textbf{40.77} \pm \textbf{10.7}$ & $\color{gray}{40.88 \pm 12.1}$ \\
        R-$2\times$ in-painting & MAE $\downarrow$ & $0.071 \pm 0.021 $ & $\textbf{0.016} \pm \textbf{0.010}$ & $0.017 \pm 0.009$ & $\color{gray}{0.005 \pm 0.002}$ \\
        R-Outpainting & MAE $\downarrow$ & $0.071 \pm 0.022$ & $0.027 \pm 0.011$ & $\textbf{0.027} \pm \textbf{0.009} $ & $\color{gray}{0.006 \pm 0.001}$ \\
        R-S\&P Denoising & MAE $\downarrow$ & $0.072 \pm 0.023$ & $\textbf{0.033} \pm \textbf{0.009}$ & $0.033 \pm 0.015$ & $\color{gray}{0.003 \pm 0.000}$ \\

        & & \multicolumn{4}{c}{\textit{Domain generalization to RETOUCH-cirrus}} \\
        R-Rec. Fluid Seg. & IoU $\uparrow$ & $37.25 \pm 10.6 $ & $\textbf{44.91} \pm \textbf{16.6} $ & $42.58 \pm 13.6$ & $\color{gray}{45.72 \pm 16.0 }$ \\
        R-$2\times$ in-painting & MAE $\downarrow$ & $0.085 \pm 0.017$ & $0.011 \pm 0.002$ & $\textbf{0.010} \pm \textbf{0.002}$ & $\color{gray}{0.005 \pm 0.001 }$ \\
        R-Outpainting & MAE $\downarrow$ & $0.085 \pm 0.017 $ & $\textbf{0.014} \pm \textbf{0.002}$ & $0.015 \pm 0.002 $ & $\color{gray}{0.007 \pm 0.001}$ \\
        R-S\&P Denoising & MAE $\downarrow$ & $0.086 \pm 0.017$ & $0.071 \pm 0.009$ & $\textbf{0.029} \pm \textbf{0.004}$ & $\color{gray}{0.001 \pm 0.000}$ \\
        
    \bottomrule
    \end{tabular}
    \label{tab:domain-generalization}
\end{table}
}

\textbf{Quantitative results}
In~\Cref{tab:unseen}, we see the results of the models when adapting to unseen OCT tasks.
First off, we see the varying difficulty of the tasks by the loose bounds set with the Copy and Single-task baselines, \eg, the binary segmentation tasks is very simple, as the copy baseline already achieves a high IoU.
Yet, we see, that only the Retinalizer model with the proposed color augmentation is able to exceed this naive baseline and coherently predict the merged retinal layers with an IoU of $81.94\%$.
Handling different color-prompts in the retinal recoloring multi fluid segmentation task is done best by the same model, which is expected, as during training, \emph{Retinalizer Rec.} has been exposed to different semantic color variations.
Here, it even surpasses the Single-task baseline, indicating that a broad pre-training on diverse tasks may in itself already have a positive effect.
Finding fine-grained structures, such as retinal boundaries is a challenging task for the models, which might hint at an architectural shortcoming, still, the Retinalizer Rec. model achieves the closest score to the Single-task model with $60.69\%$ F-1.
Moving to the generative tasks, we can see that Retinalizer with and without recoloring augmentation lead to the best results, with generally quite similar reconstruction errors.
Most of the time, the recoloring augmentation has a positive effect, merely for salt and pepper denoising it leads to slightly worse results, indicating, its main effects unfold for semantic tasks.

Moving to domain generalization experiments in~\Cref{tab:domain-generalization}, we can, for the majority of tasks confirm, that the recoloring augmentation has a positive effect and improes generalization to new data distributions.
Only in the setting when adapting to Cirrus the augmentation strategy produces slightly lower segmentation scores as compared to Retinalizer.
These results hint at bridging domain gaps via in-context learning to be a promising research direction.

\begin{figure}[t]
    \centering
    \scriptsize
    \begin{tabular}{m{2.2cm}m{1.5cm}m{1.5cm}m{1.5cm}m{1.5cm}m{1.5cm}m{1.5cm}}
         \toprule
         & \multicolumn{1}{c}{Query} & \multicolumn{1}{c}{Neuralizer} & \multicolumn{1}{c}{Retinalizer} & \multicolumn{1}{c}{Ret. Rec.}  & \multicolumn{1}{c}{Single-task} & \multicolumn{1}{c}{Target}\\
         \midrule
        R-Rec. Fluid Seg. & \includegraphics[width=0.12\textwidth,height=0.1\textwidth]{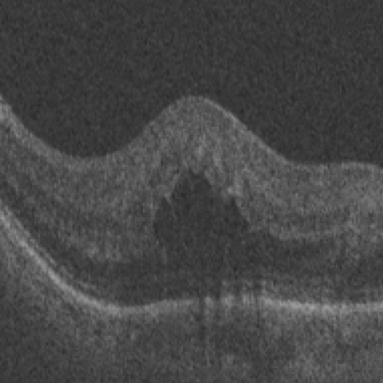} & \includegraphics[width=0.12\textwidth,height=0.1\textwidth]{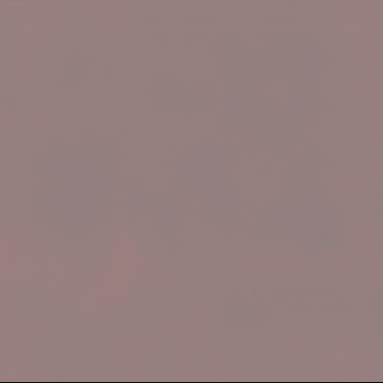} & \includegraphics[width=0.12\textwidth,height=0.1\textwidth]{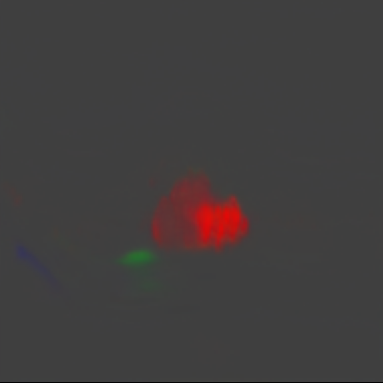} & \includegraphics[width=0.12\textwidth,height=0.1\textwidth]{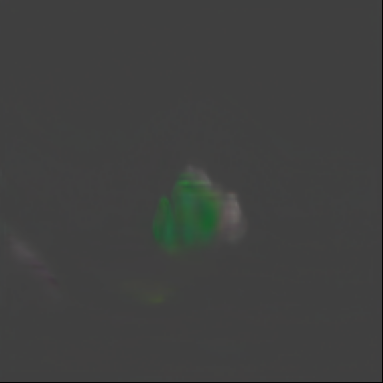} & \includegraphics[width=0.12\textwidth,height=0.1\textwidth]{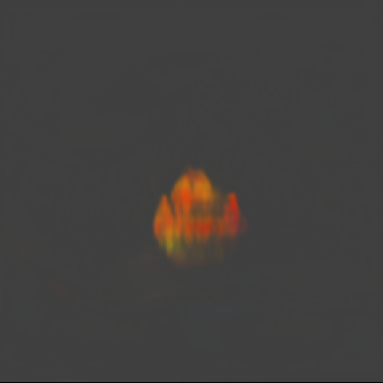} & \includegraphics[width=0.12\textwidth,height=0.1\textwidth]{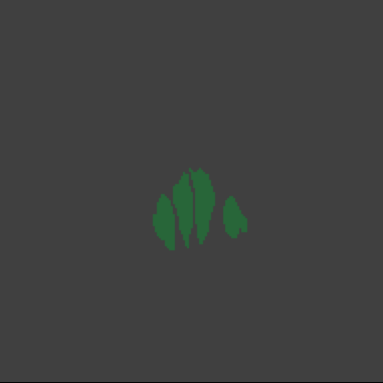} \\
        
        D-Layer Bin. Seg. & \includegraphics[width=0.12\textwidth,height=0.1\textwidth]{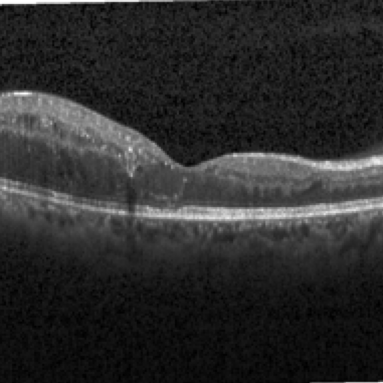} & \includegraphics[width=0.12\textwidth,height=0.1\textwidth]{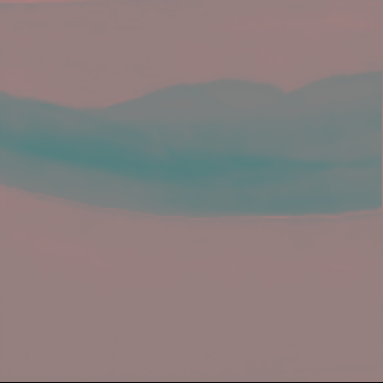} & \includegraphics[width=0.12\textwidth,height=0.1\textwidth]{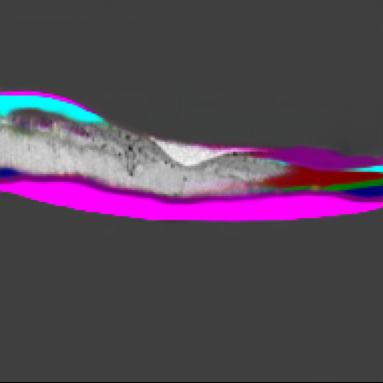} & \includegraphics[width=0.12\textwidth,height=0.1\textwidth]{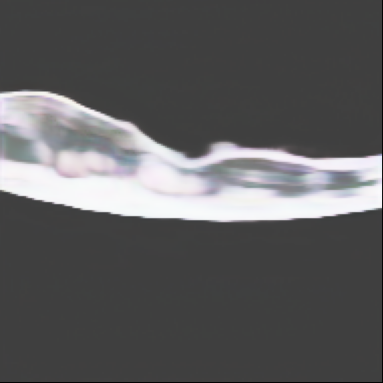} & \includegraphics[width=0.12\textwidth,height=0.1\textwidth]{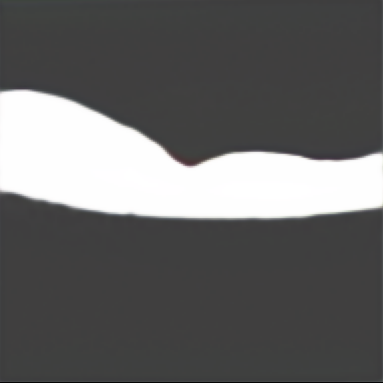} & \includegraphics[width=0.12\textwidth,height=0.1\textwidth]{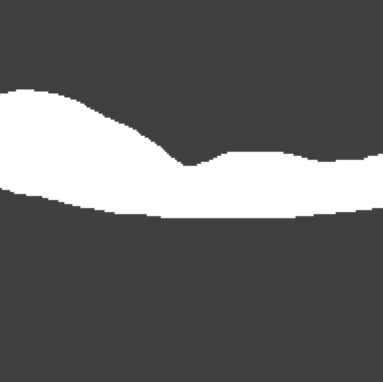} \\
        
        D-Layer Bound. & \includegraphics[width=0.12\textwidth,height=0.1\textwidth]{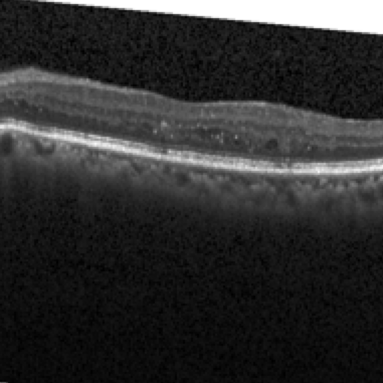} & \includegraphics[width=0.12\textwidth,height=0.1\textwidth]{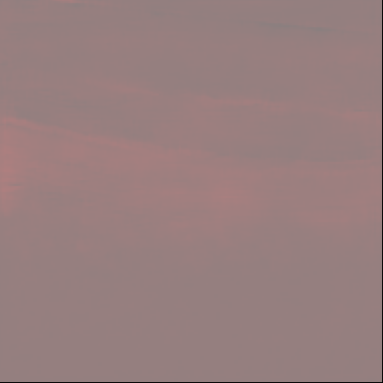} & \includegraphics[width=0.12\textwidth,height=0.1\textwidth]{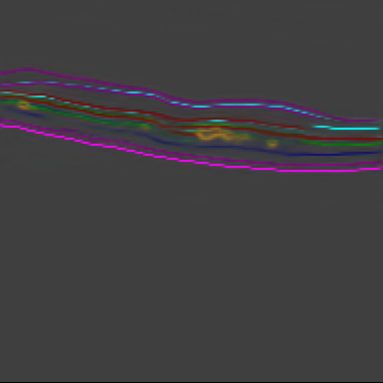} & \includegraphics[width=0.12\textwidth,height=0.1\textwidth]{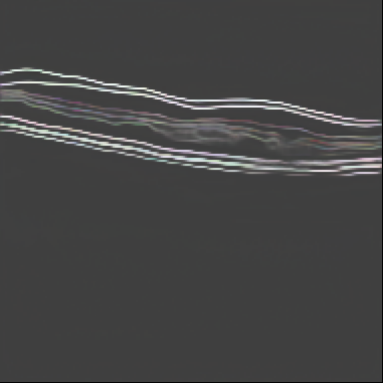} & \includegraphics[width=0.12\textwidth,height=0.1\textwidth]{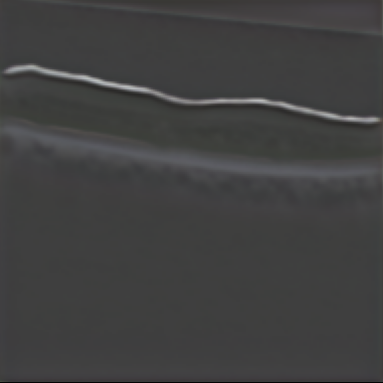} & \includegraphics[width=0.12\textwidth,height=0.1\textwidth]{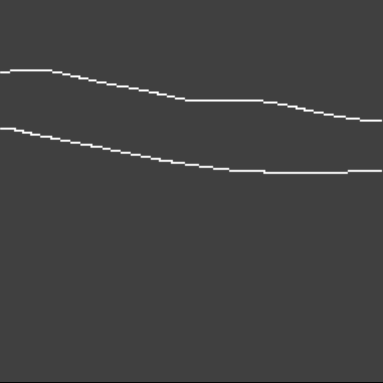} \\
        
        O-$2\times$ in-painting & \includegraphics[width=0.12\textwidth,height=0.1\textwidth]{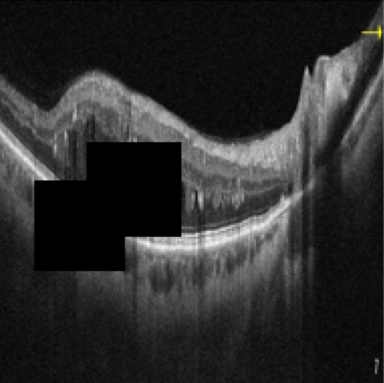} & \includegraphics[width=0.12\textwidth,height=0.1\textwidth]{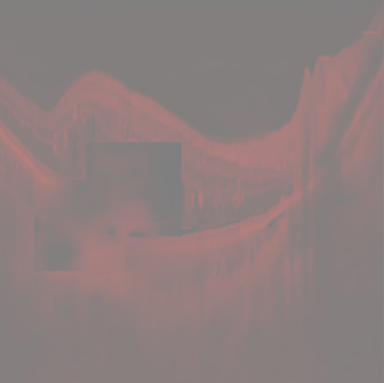} & \includegraphics[width=0.12\textwidth,height=0.1\textwidth]{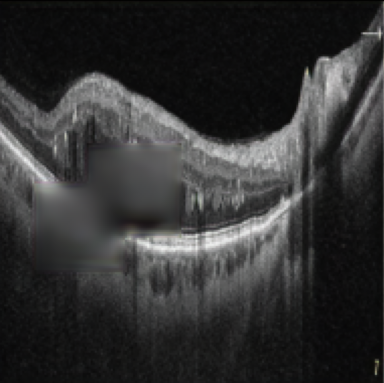} & \includegraphics[width=0.12\textwidth,height=0.1\textwidth]{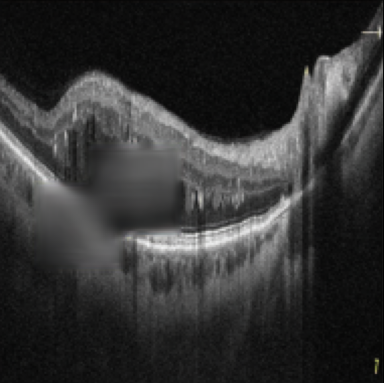} & \includegraphics[width=0.12\textwidth,height=0.1\textwidth]{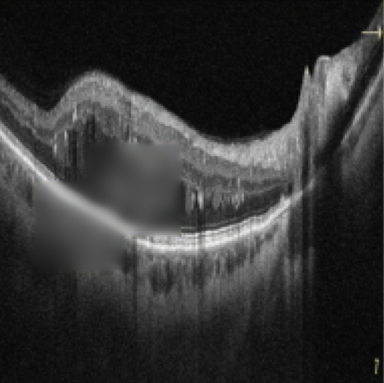} & \includegraphics[width=0.12\textwidth,height=0.1\textwidth]{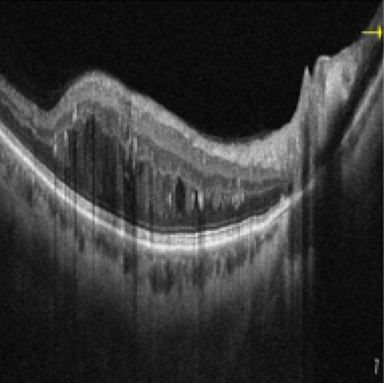} \\
         \bottomrule
    \end{tabular}
    \caption{Qualitative predictions of baselines and our Retinalizers on unseen tasks.}
    \label{fig:qualitative}
\end{figure}

\textbf{Qualitative results} In~\Cref{fig:qualitative}, we see the inference results of different trained models.
The Neuralizer model, not trained on OCTs, as expected can not be used to faithfully address tasks on the retina, further all predictions are predominantly red, as the model is originally trained to predict one channel outputs.
Our trained Retinalizer is able to predict semantic regions, which correspond to the correct regions to segment.
Yet, it tries to infer these regions by \emph{interpolating} between outputs of seen tasks, \ie, it tries to generalize through seen task interpolation.
When we add the recoloring augmentation to Retinalizer, we see, that it can much better generalize to unseen tasks, as variance in colorization of segmentation tasks was already introduced during training, enabling a swift adaptation.
Similar behavior can be seen for Layer Boundary detection.
For the reconstructive task in the bottom row of the figure, all models produce similarly washed out predictions, indicating that for generative pixel-prediction, an extension of the MSE loss we train with and further integrate an adversarial objective might lead to more faithful generated OCTs.

\section{Conclusion}
In this work, we brought the first visual in-context learning models to the domain of OCT, enabling to solve diverse and individualized retinal tasks, while overcoming the severe restrictions of small amount of available data and a small compute budget.
We showed, that simply from OCT segmentation datasets, we can train models addressing a diverse set of tasks from semantic- to generative tasks and adapt to new unseen tasks and data distributions on the fly at test time.
A limitation is, that in general the metric scores do not yet reach specialist model performance.
Thus, in future work we aim at expanding the number of pre-training tasks and build further task-augmentations beside our recoloring strategy which we believe to be key for better adaptation to unseen tasks.

\noindent\textbf{Acknowledgment}
This work was supported by funding from the pilot program Core-Informatics of the Helmholtz Association (HGF).

%
%
%

%
%
%
\bibliographystyle{splncs04}
\bibliography{bibliography}








\end{document}